\documentclass[10pt,twocolumn,letterpaper]{article}

\usepackage{iccv}              

\definecolor{iccvblue}{rgb}{0.21,0.49,0.74}
\usepackage[pagebackref,breaklinks,colorlinks,allcolors=iccvblue]{hyperref}
\usepackage[accsupp]{axessibility}
\def\paperID{8917} 
\def\confName{ICCV}
\def\confYear{2025}
\usepackage{graphicx}
\usepackage{graphicx}
\usepackage{multirow}
\usepackage{url}
\definecolor{highlight-blue}{HTML}{388AF5}
\usepackage{listings}
\title{CAP\includegraphics[width=0.75cm]{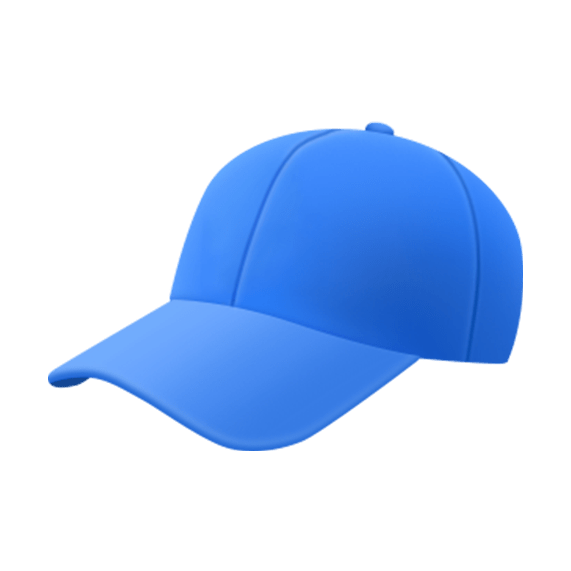} : Evaluation of Persuasive and Creative Image Generation } 

\author{Aysan Aghazadeh \\
University of Pittsburgh\\
Pittsburgh, PA\\
{\tt\small aya34@pitt.edu}
\and
Adriana Kovashka\\
University of Pittsburgh\\
Pittsburgh, PA\\
{\tt\small kovashka@cs.pitt.edu}
}

\begin{document}
\maketitle
\begin{abstract}
We address the task of advertisement image generation and introduce three evaluation metrics to assess Creativity, prompt Alignment, and Persuasiveness (CAP) in generated advertisement images. Despite recent advancements in Text-to-Image (T2I) methods and their performance in generating high-quality images for explicit descriptions, evaluating these models remains challenging. Existing evaluation methods focus largely on assessing alignment with explicit, detailed descriptions, but evaluating alignment with visually implicit prompts remains an open problem. Additionally, creativity and persuasiveness are essential qualities that enhance the effectiveness of advertisement images, yet are seldom measured. To address this, we propose three novel metrics for evaluating the creativity, alignment, and persuasiveness of generated images. We show that current T2I models struggle with creativity, persuasiveness, and alignment when the input text is implicit messages. We further introduce a simple yet effective approach to enhance T2I models' capabilities in producing images that are better aligned, more creative, and more persuasive. Code is available at \href{https://aysanaghazadeh.github.io/CAP/}{https://aysanaghazadeh.github.io/CAP/}
\end{abstract}    

\section{Introduction}
\label{sec:intro}

\begin{figure}[!tp]
    \centering
    \includegraphics[width=1\linewidth]
    {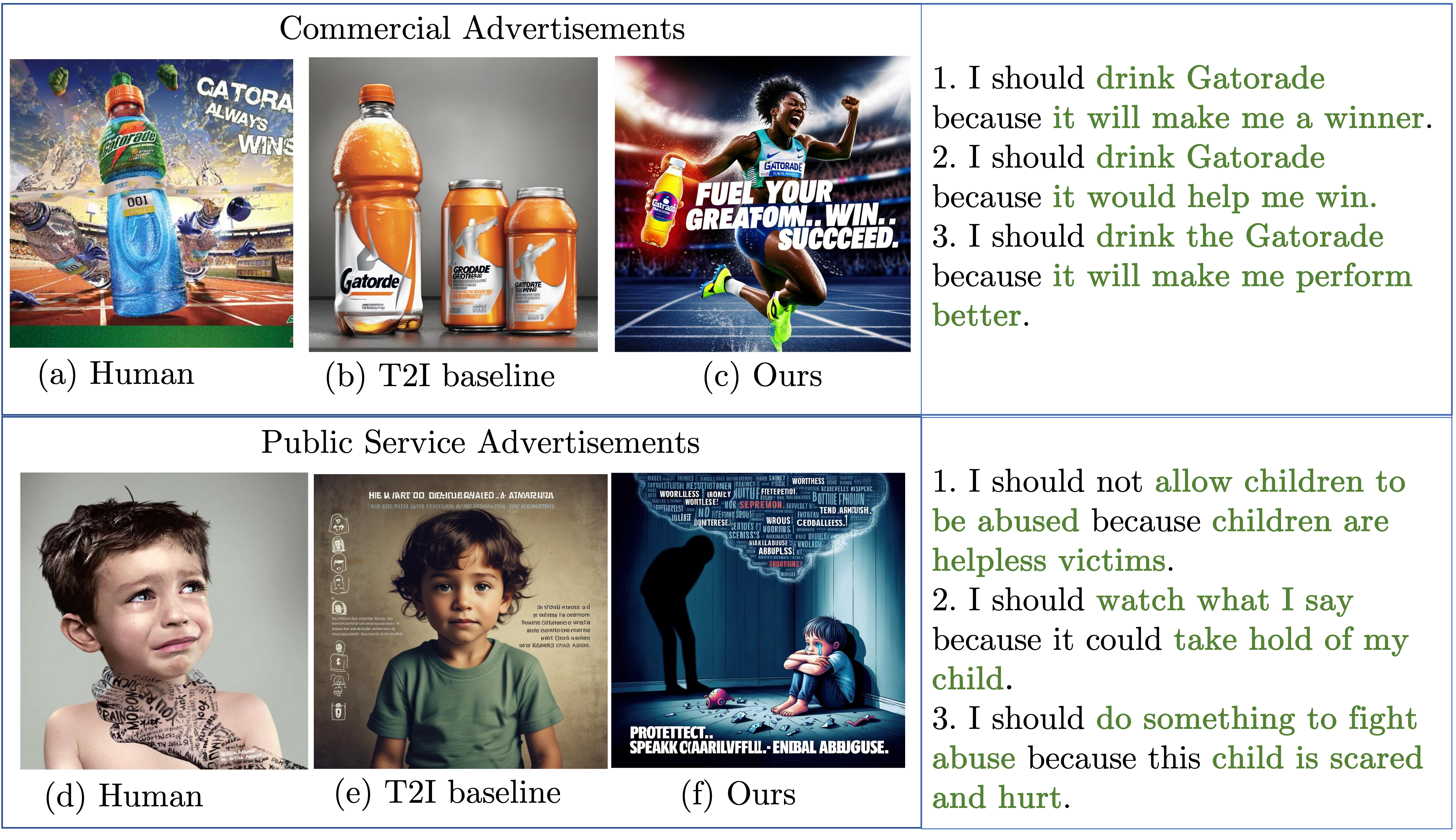}
    \caption{
   Human-created ads (a, d) convey their message through creative and persuasive visual storytelling, blending the implicit message seamlessly into the visuals.
   The T2I baseline (b, e), on the other hand, depicts relevant entities but without the underlying intent.
   This highlights the need for better metrics for \textbf{persuasiveness}, \textbf{creativity}, and \textbf{abstract text-image alignment}. Ours (c, f) demonstrates how improving visual storytelling can enhance ad generation by aligning visuals with the intended message effectively. Texts on the right form the prompt to T2I models. }
    \label{fig:intro}
\end{figure}

Advertisements (ads) impact consumer decisions and appear across various media, from billboards to social platforms. Persuasive, effective ads satisfy some or all of the following \emph{three criteria}: (1) convey their message clearly, (2) convey their message in a creative and memorable way, and (3) impact their intended audience (e.g. convince the audience to buy a product). 
Thus ads must do more than simply show a set of relevant objects. For example, in Fig.~\ref{fig:intro}, images (a), (b), (c) feature Gatorade. However, (b) only shows a bottle and two cans, while (c) includes a creative slogan and an athlete running while holding the bottle, and is more likely to persuade the audience to buy Gatorade. Similarly, on the topic of child abuse, (e) shows a child, while (f) also shows a relevant ominous shadow.



Recent advances in text-to-image (T2I) models, such as Stable Diffusion \cite{SD, sdxl} and DALLE3 \cite{DALLE3}, excel at generating high-quality photorealistic images from explicit and detailed prompts that describe objects, their attributes and relations. However, a user of a T2I method for ad generation might simply know what message they want the ad to convey, and may not have thought about the exact objects the image should depict. Thus, they may use \emph{visually implicit or abstract prompts} (\ie ones that do not precisely describe how the image should look in terms of objects). 

Further, existing T2I generation metrics, like FID \cite{FID} and CLIPScore \cite{CLIP_score}, assess image fidelity and explicit text-image alignment but do not capture qualities that are important for evaluating ads, such as creativity and persuasiveness. Our work addresses this gap by proposing the first benchmark on creative and persuasive ad generation, using three novel metrics: \underline{\textbf{C}}reativity, text-image \underline{\textbf{A}}lignment, and \underline{\textbf{P}}ersuasiveness (CAP). 
These metrics correspond to the three criteria for effective ads that we posit above.

To measure how well the ad satisfies the first criterion (conveying the message clearly), we propose 
a method to evaluate the 
\textbf{\underline{A}lignment} of  \underline{I}mage and \underline{M}essage (AIM), by (1) training an LLM to interpret the argumentation (message) in the ad image, and (2) comparing the message inferred based on the generated image, to the intended message.

To measure satisfaction of the second criterion, conveying the message in a creative and memorable way, we define \textbf{Creativity} as the divergence of the generated image from typical product images while maintaining alignment with the ad message. 
Measuring satisfaction of the third criterion (impacting the intended audience) is more complex as we must consider both the ad content and the audience, through the lens of argumentation. We define \textbf{Persuasiveness} as being convincing in driving the message of the advertisement message to the intended audience while also incorporating well-known argumentation strategies \cite{appeal, audience, creativity}.

To ground our work, we leverage \cite{PittAd}, 
which gathers annotations similar to the implicit prompts that users might employ to generate ad images. These annotations are in the form of \emph{action-reason statements}; they contain the action the ad suggests the viewer should take (e.g., ``drink Gatorade'') and the reason the ad provides for taking this action (e.g., ``it would help me win'').
We use this dataset as a source of real ads and ad action-reason messages based on which we generate additional ads. 
On this data, our proposed alignment, creativity, and persuasiveness metrics achieve higher agreement with human judgment than baseline metrics, by 0.37, 0.50, and 0.53 (out of 1).

Further, we find that public T2I models struggle to generate creative and persuasive images aligned with the action-reason statements. To address this T2I shortcoming, we propose a simple yet effective method to create expanded advertisement messages ($\mathcal{E_{LLM}}$) which use LLMs' reasoning ability to explicitly describe what visual content an ad for the corresponding messages should contain. These expanded messages are then used as prompts for the T2I models to generate ads. Fig.~\ref{fig:intro} (c), produced by this method, successfully captures both the explicit and implicit aspects of the advertising message by including a slogan and showing an athlete running and holding the bottle of Gatorade.
This simple approach improves alignment, creativity, and persuasiveness by (6\%, 7\%, 10\%) and (39\%, 38\%, 43\%) in commercial ads and public service ads, respectively. These numbers correspond to one particular configuration of MLLM, LLM and T2I methods involved, and we conduct extensive experiments with many configurations. 

To summarize our contributions:
\begin{itemize}
    \item We introduce three novel metrics for advertisement image generation: Creativity, Alignment, and Persuasion.
    \item We show that current state-of-the-art public T2I models perform poorly in terms of creativity, persuasion, and alignment when generating advertisement images from abstract text, especially for public service ads.
    \item We propose a simple yet effective approach that leverages LLMs to improve the generation of ad images.
\end{itemize}

\section{Related Work}
\label{sec:related_works}

\textbf{Text-image alignment.} With the advancement of T2I models \cite{AuraFlow, sdxl, chen2023pixart} in generating high-quality images, the evaluation of text-image alignment for explicit text descriptions has been widely explored. 
Some approaches compute \textit{image-image similarity} between the ground-truth images (real images in a dataset) and the generated images, using FID \cite{FID} and IS \cite{IS}.
These methods reward high similarity of the generated images to the ground-truth images. However, this evaluation over-focuses on low-level perceptual quality while overlooking the creativity of the images and alignment with intended messages.\footnote{We experimented with FID and IS, but obtained counterintuitive results: better, more recent models like FLUX underperformed older models like SDXL, demonstrating these metrics are not appropriate for ad image generation evaluation.} Another group of metrics computes \textit{text-image alignment} between prompt and generated images, using
CLIPScore \cite{CLIP_score} and BLIPScore \cite{BLIPscore}. However, due to the limitation of these methods in understanding compositional and complex prompts, 
these methods fail even in evaluation of images generated for complex explicit prompts \cite{VQA}. 
Some methods use \textit{LLMs and MLLMs to compute image-text alignment} \cite{VQA, T2Iscorer, evalalign} by fine-tuning an LLM or MLLM to answer questions, e.g. ``Does the image show [T2I Prompt]?''
Others \cite{QCA, VQ2, TIFA} first generate questions given the T2I prompt and use these as prompts for VLMs to score the images. 
These methods have limited usability in our setting where the prompt is a message that does not explicitly mention the visual content, and two totally visually different images can convey the same message.
Other methods \cite{ImageReward, HPSV2, socialReward, PickScore} rely on \textit{human feedback}. 
Although these models are fine-tuned with (expensive) human feedback, they rely on VLMs, which have limited capacity to capture semantics in persuasive images \cite{wacv}. 
Finally, \cite{concept,fan2024prompt} focus on the image generation of concepts like peace rather than conveying a persuasive message and \cite{wu2024imagine}
proposes a method to generate images based on abstract description which still includes the visual elements. 
We propose a method to evaluate generation based on visually implicit text messages (few objects specified).

\textbf{Non-computational analysis of persuasion and creativity.}
Theories of persuasion date to Aristotle's pathos, ethos and logos \cite{braet1992ethos}. Strategies specifically for decoding ads were proposed in \cite{williamson1978decoding}.
Others \cite{appeal,braet1992ethos,pelclova2010persuasive, safitri2013persuasive} analyze persuasion strategies and their influence on effectiveness and impact on the audience \cite{audience}. 
Finally, 
\cite{shen2021influence, amad2022advertisement,creativity} analyze the influence of creativity in ads but do not propose computational metrics; our work addresses this gap.
 
\textbf{Computational analysis of persuasion.} 
Generation and evaluation of persuasive text-only content has been explored in the field of NLP. With the development of LLMs, many works  \cite{elaraby2024persuasiveness, voelkel2023artificial, palmer2023large} compare the persuasion in LLM-generated content with contents written by humans. 
\cite{rescala2024can} evaluates the performance of GPT-4 in understanding if the text content is convincing. 
\cite{singh2024measuring} introduce benchmarks on persuasiveness of texts.
\cite{zeng2024johnny, fashion} propose methods to convert a text into a persuasive one or generate a persuasive text for specific prompt. \cite{liu2022imagearg} introduce a dataset on multimodal argumentative content. \cite{PittAd} introduce a dataset of advertisements which aim to be persuasive.
\cite{guo2021detecting,kumar2023persuasion} introduce methods and benchmarks for detecting image persuasion strategies like emotion, and reasoning in advertisements. 
However, to the best of our knowledge we are the first to introduce a computational method to \emph{score the level of persuasiveness} of advertisement images, i.e. not \emph{classify the types} of strategies it uses to try to be persuasive, but \emph{whether} it is persuasive and creative. There are very few works focusing on generation of persuasive images; for example, \cite{persuasiveface} is limited to faces in ads.

\section{Methodology}
\label{sec:methods}

We introduce three novel metrics for evaluating the \textbf{\underline{C}reativity}, \textbf{\underline{A}lignment} with abstract text, and \textbf{\underline{P}ersuasiveness} (CAP) of generated images. These metrics are designed to provide a comprehensive assessment of advertisement images based on their performance in conveying visually implicit messages. Following this, we describe our simple yet effective proposed approach to address limitations in current T2I models, specifically in their performance in generation of creative and persuasive images that align with intended implicit messages. 

\subsection{Background: Task Definition} 

We define advertisement generation as the creation of creative and persuasive advertisement images $(I_{gen})$ based on implicit input messages. The messages ($AR_m$) and corresponding ground-truth images ($I_{real}$) are sourced from the PittAd dataset \cite{PittAd}. Each image in PittAd includes three to five action-reason messages ($AR_m)$, which are the interpretation of the images structured as \textit{``I should \{\textbf{action}\}$(A_m)$, because \{\textbf{reason}\}} $(R_m)$". These are combined and used as prompts to T2I models.

\begin{figure}
    \centering
    \includegraphics[width=1\linewidth]{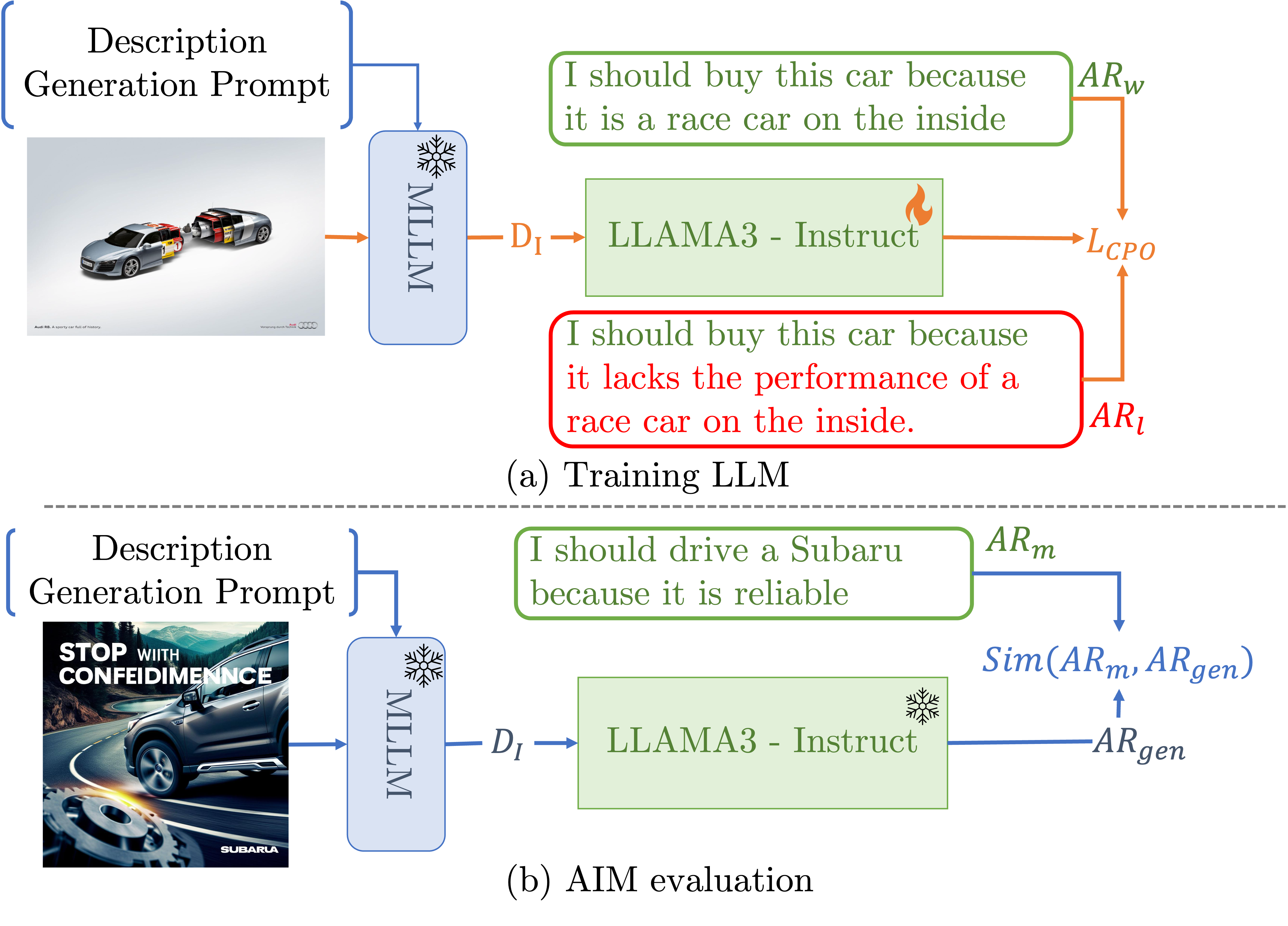}
    \caption{Overview of AIM. \textcolor{orange}{Orange} denotes training, while \textcolor{blue}{blue} is inference. $AR_w$ and $AR_l$ are used in training as the preferred and dis-preferred statements. $AR_m$ is the prompt for the T2I model.}
    \label{fig:CITE}
\end{figure}

\subsection{Alignment of Image with Abstract Text} 

In the advertisement image generation task, the input text often implies rather than explicitly describing the objects and their attributes. In these cases, generated images may appear visually relevant but fail to represent the deeper semantics or messages of the ad. For example, in Fig. \ref{fig:intro} (b), the image includes a bottle of Gatorade, aligning visually with the input text, yet it fails to capture the emphasis on the reason ``it would help me win". Existing alignment evaluation methods \cite{CLIP_score, ImageReward, VQA, VQ2, socialReward} fail to detect mismatches between messages ($AR_m$) used as prompts, and generated images, due to the abstractness (visual implicitness) of the messages. To address this, we introduce 
\textbf{A}lignment of \textbf{I}mage and \textbf{M}essage (AIM).
We utilize two steps, shown in Fig.~\ref{fig:CITE}. First, we train (fine-tune) an LLM (\eg LLAMA3-Instruct-8B) to match a description of the ad image to the correct action-reason statement, ensuring the LLM pays fine-grained attention to the semantics of the message. 
Second, we use this trained LLM to produce an action-reason statement for a generated image ($AR_{gen}$), and compute the alignment (similarity) between this produced action-reason statement and the input action-reason statement in the prompt ($AR_m$), as the AIM score for the generated image.  

In more detail, we utilize a Multimodal Large Language Model (\eg InternVL-V2-26B \cite{InternVL}) to generate the image description \textcolor{black}{$D_I$}. 
Next, we input this description into a fine-tuned LLM (\eg LLAMA3-Instruct-8B \cite{dubey2024llama}) to generate an action-reason statement for the input image (\textcolor{black}{$AR_{gen}$}).
We separately compute the similarity of the action and reason components of the $AR_m$ (\ie $A_m$ and $R_m$) with the corresponding component in the $AR_{gen}$ (in order to avoid the more visually explicit action from dominating the score). We use a semantical similarity score \cite{sim_score} and then return their weighted average as the overall alignment score:
\begin{equation}
\small
    AIM(I_{gen}, AR_m) = \frac{Sim(A_{gen}, A_m) + \alpha\ Sim(R_{gen}, R_m)}{1 + \alpha}
\end{equation}
where $\alpha$ is the weight for reason when computing the similarity between the statements, which is set\footnote{Agreement between human annotators and AIM scores computed with different $\alpha$, evaluated on a subset of images, are: $\alpha = 1: 0.51$, $\alpha = 2, 3: 0.55$, $\alpha = 4,5: 0.68$. We include an example in the supplementary file.} to 4.

We opt to fine-tune the LLM 
because it sometimes fails to generate statements with the correct \emph{message} given an image. For example, the LLAMA3-Instruct-8B output for Fig. \ref{fig:CITE} (b) is \textit{I should drive with confidence because Subaru promises a safe and reliable experience with its vehicles.} While this is an acceptable output, \emph{driving} with confidence (possibly over-confidently) is not what the image is advertising (being able to safely \emph{stop} with confidence). 

To generate more accurate interpretations for the described images, and thus enable accurate alignment predictions, we fine-tune an LLM 
with Contrastive Preference Optimization (CPO) \cite{CPO}. In CPO, each data point includes a prompt, preferred answer, and dis-preferred answer. The objective is to optimize the LLM to generate responses closer to the preferred answers while staying more distant from the dis-preferred answers. The prompts are image descriptions $(D_I)$ for 250 images from \cite{PittAd} (generated using InternVL-V2-26B \cite{InternVL}).
Each ground-truth action-reason statement from \cite{PittAd} serves as a preferred action-reason statement $(AR_w)$.
For dis-preferred action-reason statements $(AR_l)$, we used the hard negative statements from \cite{wacv}. These hard negatives share the visual elements with $AR_m$; however, the statements are semantically different. For example, in Fig. \ref{fig:CITE}, ``\textit{I should buy this car because it is a race car on the inside}" is converted to ``\textit{I should buy this car because it lacks the performance of the race car on the inside}". While the visual elements---\textit{car} and \textit{race car}---are the same in both sentences, the semantics of the sentence are completely changed. 
This process generated a dataset in size of 3500 data points. Finally, we fine-tuned the LLM using CPO on $(D_I, AR_w, AR_l)$, as shown in Fig.~\ref{fig:CITE}. 
Recall that after generation of $AR_{gen}$ by the fine-tuned LLM, we compute the similarity \cite{sim_score} between the generated action-reason statement $(AR_{gen})$ and the input message $(AR_m)$.

\subsection{Creativity} 

We define creativity in the context of advertisement image generation as (1) uniqueness, i.e. being distant from an image which simply shows the objects in the message, while maintaining (2) relevance, i.e. conveying the intended action-reason message. We posit that the probability of the image being unique should increase when it includes additional visual elements not explicitly mentioned in the prompt. 
Thus, we introduce a creativity score with two components: (1) cosine similarity between the CLIP features of the image and the objects in the text, $Sim(I_{gen}, obj$) (expected to be low for unique images); and
(2) text-image alignment between the prompt/message and image (expected to be high for relevant images). We formulate the creativity score ($C_{obj}$) as follows:
\begin{equation}
\small    
C_{obj} = \frac{AIM(I_{gen}, AR_m)}{\frac{1}{n} \times \Sigma_{obj \in objects} ~ Sim(I_{gen}, obj) + 0.01}
\end{equation}
where $n$ is the number of objects mentioned in the text, and 0.01 is a smoothing value to avoid diving by 0. The smoothing value is chosen to be small compared to average CLIP scores (which is about 0.30 in our experiments) to limit the effect of soothing value. We use an LLM to extract a list of objects mentioned in the action-reason text.


\begin{figure}
    \centering
    \includegraphics[width=1\linewidth]{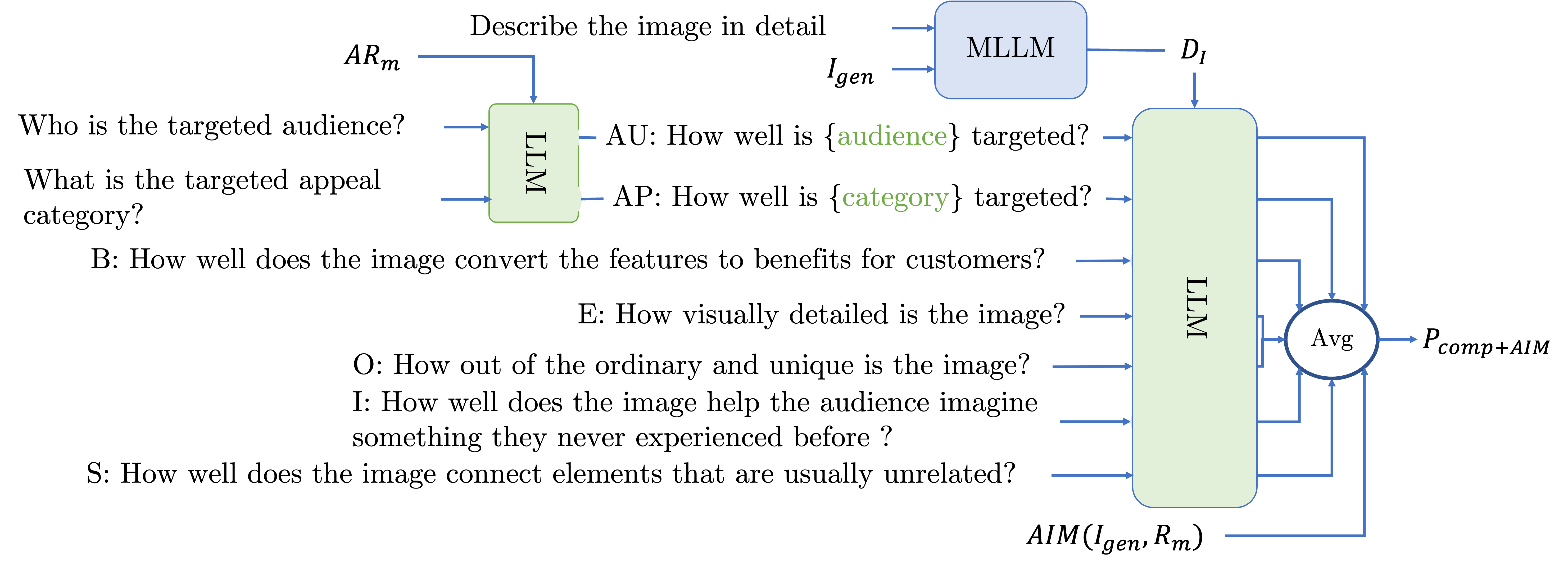}
    \vspace{-0.2cm}
    \caption{Process of computing $P_{comp+AIM}$ persuasiveness score}
    \label{fig:PA}
\end{figure}

\subsection{Persuasiveness} 

Evaluating whether an image is persuasive involves scoring the persuasiveness techniques the image uses, as well as their effect on the intended audience. To capture the complexity of the task, we propose a persuasiveness score $P_{comp}$ which relies on multiple \textit{comp}onents. These components are well grounded in prior non-computational literature on persuasiveness \cite{braet1992ethos,appeal,creativity,audience}. Some evaluate successful use of persuasiveness techniques, while others examine effect on the audience. We also incorporate our AIM score, resulting in $P_{comp+AIM}$. The components are: 
\begin{itemize}
    \item Audience (AU): Each advertisement is supposed to target specific groups of audiences. How well does the generated \emph{image} target the audience the \emph{message} targets?
    \item Benefit (B): One way to make the ads more persuasive is to depict the benefits of the product features for the customer. How well does the image convert the features to benefits for customers?
    \item Appeal Category (AP): Each advertisement is intended to target one of three appeal categories: Ethos, Pathos, or Logos. Ethos appeals to ethics and credibility; for example, when a respected celebrity endorses a product. Logos appeals to logic, using reasoning or factual information to persuade. Pathos aims to evoke an emotional response in the audience. How well does the image target the appeal category of the advertisement message?
    \item Elaboration (E): How visually detailed is the image?
    \item Originality (O): How out of the ordinary and unique is it?
    \item Imagination (I): How well does the image help the audience imagine something they never experienced?
    \item Synthesis (S): How well does the image connect elements that are usually unrelated?
\end{itemize}

The evaluation of the components is shown in Fig.~\ref{fig:PA}. We first generate a description of the image using an MLLM. Because of the superiority of LLMs in reasoning \cite{you2023idealgpt,verbalizing_emnlp,wacv}, we prompt an LLM (rather than MLLM) with each component question. We also include as input to this LLM: (1) the description of the image, and (2) if applicable, the targeted audience and the appeal category for the ad message (detected by LLM). Then, for each ``How/How well'' question (Fig.~\ref{fig:PA}), we ask the LLM to score each image in the range (0, 5). We compute $P_{comp}$  as the average of all the previous scores divided by 5. For $P_{comp+AIM}$, we also compute the alignment of the image with the intended reason $R_m$. In the supplementary file, we show the prompts for each question, and demonstrate through examples the benefit of combining the components to achieve better agreement with human scores. Combining the components acts like an ensemble model where slight inaccuracies in terms of individual components counter-balance each other. It can also be seen as accumulating evidence for persuasiveness from diverse sources for a more accurate result.

\subsection{Text-to-Image Approach} 
Our results show T2I models lack creativity and persuasiveness and fail to generate images aligned with text when the input is not an explicit description. To address these issues in  existing models, we propose to first generate an expanded message for an advertisement image $(\mathcal{E}_{LLM})$ utilizing an LLM. We prompt the LLM with ``Describe an advertisement image that conveys the following messages in detail: [$AR_m$]. Only return one paragraph of description without further explanation." Next, we use the resulting expansion as the prompt for the T2I method. Thus instead of prompting the T2I model with an implicit message, we use the creativity of LLMs to generate a detailed, visually explicit description, and use it as the prompt for T2I. We refer to the resulting images as $I_{LLM}$. In contrast, the images directly generated with the action-reason prompt are denoted $I_{AR}$.

\section{Experimental Setup}
\label{sec:experimental_setup}

\textbf{Dataset.} We utilize the PittAd \cite{PittAd} dataset. Each image has three to five interpretations ($AR_m$) that include an action component ($A_m$) and a reason component ($R_m$). Each image in this dataset has one or more topic annotations. We use these to categorize advertisement messages into two groups: 1. Commercial Advertisements: Covering topics like Cars, Fashion, etc. 2. PSA Advertisements: Covering topics like Human Rights, Animal Rights, etc. We only consider topics labeled in an image by at least two annotators. To ensure balance, consistency in experiments and accuracy in comparisons between commercial and PSA advertisements, we sampled 250 PSA advertisements (the total number of images that matched our criteria for PSA advertisements in the test size of 4000 from \cite{wacv}) and 300 commercial advertisements from the test set.

\textbf{Baseline Metrics.} To the best of our knowledge, this is the first work proposing computational metrics for evaluating creativity and persuasiveness. Given the superior understanding of ads by LLMs compared to MLLMs \cite{you2023idealgpt,verbalizing_emnlp,wacv}, we use direct queries to an LLM ($C_{LLM}$ and $P_{LLM}$) as our baselines. The LLM (\eg LLAMA3-Instruct-8B \cite{dubey2024llama}) is given a description of the image from an MLLM (\eg InternVL-V2-26B \cite{InternVL}). For alignment evaluation, we compare our proposed metric, AIM, with three state-of-the-art text-to-image evaluation metrics: VQAScore \cite{VQA}, ImageReward \cite{ImageReward}, and CLIPScore \cite{CLIP_score}.
In VQAScore, an LLM is fine-tuned to answer the question, ``Does this figure show [T2I prompt]?" The model outputs the probability of ``yes" as the alignment score for the image. Following the paper's recommendations, we used the CLIP-FlanT5 version of VQAScore. ImageReward uses (expensive) human feedback on images to fine-tune its model. 
We also compare our proposed AIM metric with a zero-shot version of AIM (non-fine-tuned LLM) to show the importance of fine-tuning.

\textbf{LLM Models.} We test different LLMs for (1) evaluation (computing AIM), and (2) producing the expansion $(\mathcal{E}_{LLM})$. We use LLAMA3-Instruct-8B (LLAMA3) \cite{dubey2024llama} and Qwen2.5-7B-Instruct (QwenLM) \cite{QWenLM}.

\textbf{MLLM Models.} We experiment with InternVL2-26B (InternVL) \cite{InternVL} and Qwen2-VL-7B-Instruct (QwenVL) \cite{QWenVL} to produce the image description. 

\textbf{T2I Models.}  Since prompts generated by LLMs ($\mathcal{E}_{LLM}$) are often longer and more detailed, the effectiveness of the text tokenizer in the T2I model influences performance. After evaluating various T2I models, we selected AuraFlow-v0.2 (AuraFlow)  \cite{AuraFlow} and Flux-v1-dev (Flux) \cite{Flux} for both $I_{LLM}$ (our proposed expansion method) and $I_{AR}$ (directly inputting action-reason to T2I) due to its ability to handle long prompts effectively. sdxl-flash (SDXL) \cite{sdxl} was additionally chosen for $I_{AR}$ generation (but performed poorly with $I_{LLM}$ due to its text tokenizer).

\textbf{Human Annotation.} To assess the accuracy of our proposed metrics, we selected 60 pairs of generated images (120 total) and six pairs of real images. Each pair of generated images includes one $I_{AR}$ and one $I_{LLM}$ image (created with LLM expansion of the AR message), in random order. 
Each pair was evaluated by 2 annotators (56 unique annotators overall) asked to rank the images based on these criteria: 
(1) Alignment with action ($A_m$) and reason ($R_m$), (2) Persuasiveness, (3) Creativity, (4) Targeting of the correct audience, (5) Targeting correct appeal category, (6) Effective conversion of features into customer benefits, (7) Originality, (7) Imagination, (8) Elaboration, and (9) Synthesis. 
The annotation interface is included in the supplementary file.
We report Krippendorff’s Alpha \cite{krippendorff2011computing} to measure agreement between annotators and different metrics. To evaluate agreement with AIM, we identify the image most frequently selected by annotators, across questions on this pair about alignment with the multiple $A_m$ and $R_m$. For creativity, we report the agreement between the ranking of images based on $C_{obj}$ and the annotators' creativity rankings. For persuasiveness, we assess the agreement between the ranking based on $P_{comp+AIM}$ and the annotators' most frequently chosen image. We also analyze agreement across each component used to calculate persuasiveness.

\textbf{Implementation.} 
We fine-tuned LLAMA3 and QwenLM using CPO \cite{CPO}, with a batch size of 4, learning rate of 5e-5, and 3,000 training steps. 
All experiments were conducted on A100 and L40s GPUs.
Other details in supp.

\section{Results}
\label{sec:results}

This section evaluates the advertisement images based on their alignment with abstract messages, and their creativity and persuasiveness. We showcase the superiority of our proposed metrics---AIM, $C_{obj}$, and $P_{comp+AIM}$---by comparing them to baseline metrics using human agreement. We also show the benefits of our strategy for text-to-image generation: expanding the message with an LLM ($I_{LLM}$). 

\subsection{Alignment}

\begin{table}[!tp]
\centering
\scriptsize
\setlength{\tabcolsep}{0.9pt}

\begin{tabular}{l|c|c|c}
    \toprule  
    Annotators & COM & PSA  & All \\
    \midrule
    \midrule
    H,  ImageReward & 0.12& 0.06 & 0.11 \\
    H, VQAScore & 0.38 & 0.24 & 0.31 \\
    H, CLIPScore & 0.04 & 0.34 & 0.17\\
    H, AIM (InternVL, LLAMA3) (0-shot) & 0.17 & 0.26 & 0.18 \\
    H, AIM (InternVL, LLAMA3) &0.60 & \textbf{0.82} & \textbf{0.68} \\
    H, AIM (InternVL, QwenLM) & 0.62 & 0.56 & 0.60\\
    H, AIM (QwenVL, LLAMA3) & 0.62 & 0.56 & 0.60 \\
    H, AIM (QwenVL, QwenLM) & \textbf{0.72} & 0.56 & 0.65\\
    \hline
    H1, H2 & 0.86 &  0.85 & 0.86 \\

\bottomrule
\end{tabular}
\caption{Agreement among human annotators (H) and alignment AI metrics. We show the impact of different MLLMs and LLMs for AIM (Fig.~\ref{fig:CITE}) by ([MLLM], [LLM]).
H1, H2 are the annotators. Highest H-AI scores bolded.}
\label{tab:alignment_agreement}
\end{table}

\begin{table}[!tp]
\centering
\setlength{\tabcolsep}{0.8pt}
\scriptsize
\begin{tabular}{l|l|l|l|c|c|c|c|c|c}
\hline
\multicolumn{2}{c|}{\textbf{Evaluation}} & \multicolumn{2}{c|}{\textbf{Image}} &\multicolumn{3}{c|}{\textbf{COM Ads}} & \multicolumn{3}{c}{\textbf{PSA Ads}} \\ \cline{1-10} 
MLLM & LLM & \textbf{T2I} &  \textbf{$I_{input-text}$}& \textbf{AIM} & \textbf{$C_{obj}$} & \textbf{$P_{c+A}$} & \textbf{AIM} & \textbf{$C_{obj}$} & \textbf{$P_{c+A}$} \\ \hline
\hline
\multirow{14}{*}{InternVL}& \multirow{7}{*}{LLAMA3}& SDXL & $I_{AR}$ & 0.50 & 2.03 & 0.62 & 0.32 & 1.33 & 0.48\\ \cline{3-10}
& & \multirow{3}{*}{AuraFlow} & $I_{AR}$    & 0.50 &  2.12 & 0.64                   & 0.31  &  1.36  & 0.42\\
& & & $I_{LLAMA3}$                          & 0.53 &  2.25  & \textbf{0.70}         & 0.43 &  1.87 & \textbf{0.60}\\
& & & $I_{QwenLM}$                          & \textbf{0.55} & \textbf{2.34} & 0.59  & \textbf{0.48} &  \textbf{2.05} & 0.54\\
\cline{3-10}
& & \multirow{3}{*}{FLUX} & $I_{AR}$    & 0.51 &  2.06   & 0.48                 & 0.43  &  1.83  & 0.44\\
& & & $I_{LLAMA3}$                      & \textbf{0.54} &  2.20  & 0.52         & \textbf{0.47} &  1.93 & \textbf{0.53}\\
& & & $I_{QwenLM}$                      & 0.53 & \textbf{2.35} & \textbf{0.55}  & \textbf{0.47} &  \textbf{2.06} & 0.46\\
\cline{2-10}
& \multirow{7}{*}{QwenLM}& SDXL & $I_{AR}$ & 0.49 & 1.90 & 0.50 & 0.44 & 1.32 & 0.35\\ 
\cline{3-10}
& & \multirow{3}{*}{AuraFlow} & $I_{AR}$    & 0.47 &  1.91   & \textbf{0.50}            & 0.30  &  1.28  & 0.41\\
& & & $I_{LLAMA3}$                          & 0.50 &  \textbf{2.50}  & \textbf{0.50}    & \textbf{0.49} &  1.83 & 0.44\\ 
& & & $I_{QwenLM}$                          & \textbf{0.51} &  2.18  & \textbf{0.50}    & 0.48 & \textbf{2.04} & \textbf{0.45}\\
\cline{3-10}
& & \multirow{3}{*}{FLUX} & $I_{AR}$    & 0.50 &  1.94   & 0.49                         & 0.47  &  1.86  & 0.45\\
& & & $I_{LLAMA3}$                      & \textbf{0.51} &  2.04  & 0.49                 & \textbf{0.48} &  1.93 & 0.45\\
& & & $I_{QwenLM}$                      & \textbf{0.51} & \textbf{2.18} & \textbf{0.56} & \textbf{0.48} &  \textbf{2.03} & \textbf{0.46} \\
\cline{1-10}
\multirow{14}{*}{QwenVL}& \multirow{7}{*}{LLAMA3} & SDXL & $I_{AR}$ & 0.52 & 2.06 & 0.52 & 0.45 & 1.88 & 0.38\\ 
\cline{3-10}
& & \multirow{3}{*}{AuraFlow} & $I_{AR}$    & 0.51 &  2.08   & 0.44                             & 0.45 &  1.92  & 0.35\\
& & & $I_{LLAMA3}$                          & 0.53 &  2.19  & \textbf{0.54}                     & 0.47 &  2.03 & 0.46\\
& & & $I_{QwenLM}$                          & \textbf{0.54} & \textbf{2.32} & \textbf{0.54}     & \textbf{0.48} & \textbf{2.08} & \textbf{0.47}\\
\cline{3-10}
& & \multirow{3}{*}{FLUX} & $I_{AR}$    & 0.51 &  2.02 & 0.47                   & 0.46  &  1.89  & 0.42\\
& & & $I_{LLAMA3}$                      & \textbf{0.53} & 2.17  & \textbf{0.49} & 0.47 &  1.97 & 0.45\\
& & & $I_{QwenLM}$                      & \textbf{0.53} & \textbf{2.30} & 0.47  & \textbf{0.48} & \textbf{2.09} & \textbf{0.46}\\
\cline{2-10}
& \multirow{7}{*}{QwenLM}& SDXL & $I_{AR}$ & 0.49 & 1.93 & 0.43 & 0.44 & 1.84 & 0.37\\ 
\cline{3-10}
& & \multirow{3}{*}{AuraFlow} & $I_{AR}$    & 0.48 &  1.95  & 0.43                              & 0.44  &  1.87  & 0.36\\
& & & $I_{LLAMA3}$                          & 0.50 &  2.18  & 0.46                              & 0.46 &  1.97 & 0.43\\
& & & $I_{QwenLM}$                          & \textbf{0.52} & \textbf{2.20} & \textbf{0.47}     & \textbf{0.48} & \textbf{2.05} & \textbf{0.44}\\
\cline{3-10}
& & \multirow{3}{*}{FLUX} & $I_{AR}$    & 0.48 &  1.94   & 0.44                             & 0.46 & 1.92 & 0.40\\
& & & $I_{LLAMA3}$                      & 0.49 & 2.03  & \textbf{0.47}                      & 0.47 &  1.97 & 0.43\\
& & & $I_{QwenLM}$                      & \textbf{0.52} & \textbf{2.20} & \textbf{0.47}     & \textbf{0.48} & \textbf{2.06} & \textbf{0.44}\\
\hline
\end{tabular}
\caption{Comparison of text-image alignment (AIM), $C_{obj}$, and persuasiveness ($P_{c+A}$, short for $P_{comp+AIM}$). Best result per group is bolded. MLLM is the image descriptor. LLM is used in computing AIM and $P_{c+A}$ (and separately for input text).}
\label{tab:creativity_alignment}
\end{table}

\textbf{Baseline alignment metrics struggle with implicit input text.} 
Table \ref{tab:alignment_agreement} - column All, which is the agreement over all the images in our human evaluation, shows that the agreement between human annotators in evaluation of alignment is 0.86. We show variants of the AIM metric using different MLLMs used for description (Fig.~\ref{fig:CITE}) and LLMs. The best agreement between human annotators and the AIM metric is 0.68 (substantial agreement \cite{agreement}). In contrast, agreement with ImageReward is 0.11 (none to slight agreement \cite{agreement}), with VQAScore is 0.31 (fair agreement \cite{agreement}), and with CLIPScore is 0.17 (none to slight agreement), across all images. The gap in agreement scores between our proposed metric (AIM) and baseline metrics is even more significant for PSA advertisements (\eg compared to VQAScore). Unlike commercial ads, which often contain recognizable objects (\eg the Gatorade in Fig.~\ref{fig:intro}), PSAs typically do not promote a specific product hence the exact content to be depicted is even more implicit. This larger gap in PSA evaluation demonstrates the limitations of baseline metrics in effectively assessing images generated from implicit text prompts. 
This table also demonstrates the impact of training the LLM (compared to the zero-shot version).
Fig.~\ref{fig:example} shows examples of the alignment by baseline and proposed metrics.
In the supplementary, we also include results from scoring ads with baseline metrics (VQAScore, CLIPScore, ImageReward) on both real and generated images. We observe that some baseline metrics score \emph{real} ad images \emph{worse} than generated ones, which contradicts expectations and calls these metrics' reliability into question.

\begin{figure*}
    \centering
    \includegraphics[width=0.95\linewidth]{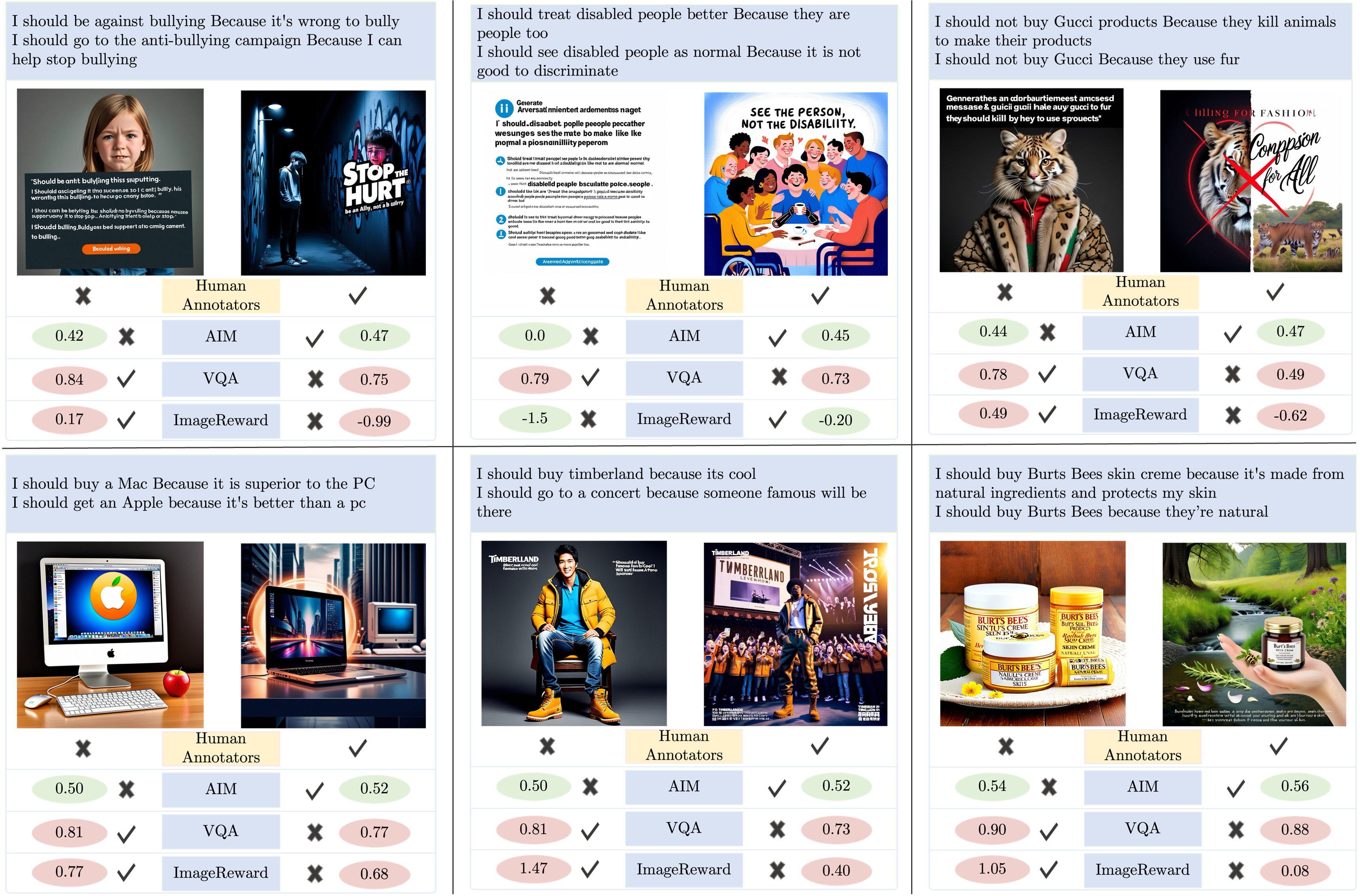}
    \vspace{-0.2cm}
    \caption{Example of images chosen by each annotator between $I_{AR}$ (left) and $I_{LLAMA3}$ (right). For each pair of images, annotators select the image that better aligns with each $AR_m$. 
    In each row, the value under each image indicates the score generated by the metric listed. A $\checkmark$ represents the chosen (ranked better) image, while a $\times$ indicates the rejected image. The \textcolor{Green}{green} circle highlights agreement with human annotations in choosing the better-aligned image, and the \textcolor{red}{red} circle indicates disagreement.}
    \label{fig:example}
\end{figure*}

\textbf{T2I models struggle with implicit text inputs in image generation.} 
In Table~\ref{tab:creativity_alignment}, we compare $I_{AR}$ and $I_{LLM}$ using different choices of: (1) MLLM for image description, (2) LLMs for computing AIM, (3) text-to-image models, and (4) LLMs for message expansion (resulting in different $I_{LLM}$). The table shows that using a simple approach of first generating descriptions with an LLM for what an ad with this message should contain, and then using the description as T2I input, improves the alignment of the generated images with $AR_m$. 
Within each group of results ($I_{AR}$, $I_{LLAMA3}$ and $I_{QwenLM}$), the best AIM result is always from one of the LLMs. In most cases, $I_{QwenLM}$ outperforms $I_{LLAMA3}$.
Results are not significantly different using the different MLLMs (InternVL and QwenVL) and for different evaluation LLMs (LLAMA3 and QwenLM). T2I methods AuraFlow and FLUX also have similar performance, with FLUX having an edge in some cases (e.g. PSAs using $I_{AR}$ with InternVL for MLLM).

\subsection{Creativity}


\begin{figure}
    \centering
    \vspace{-0.4cm}
    \includegraphics[width=0.7\linewidth]{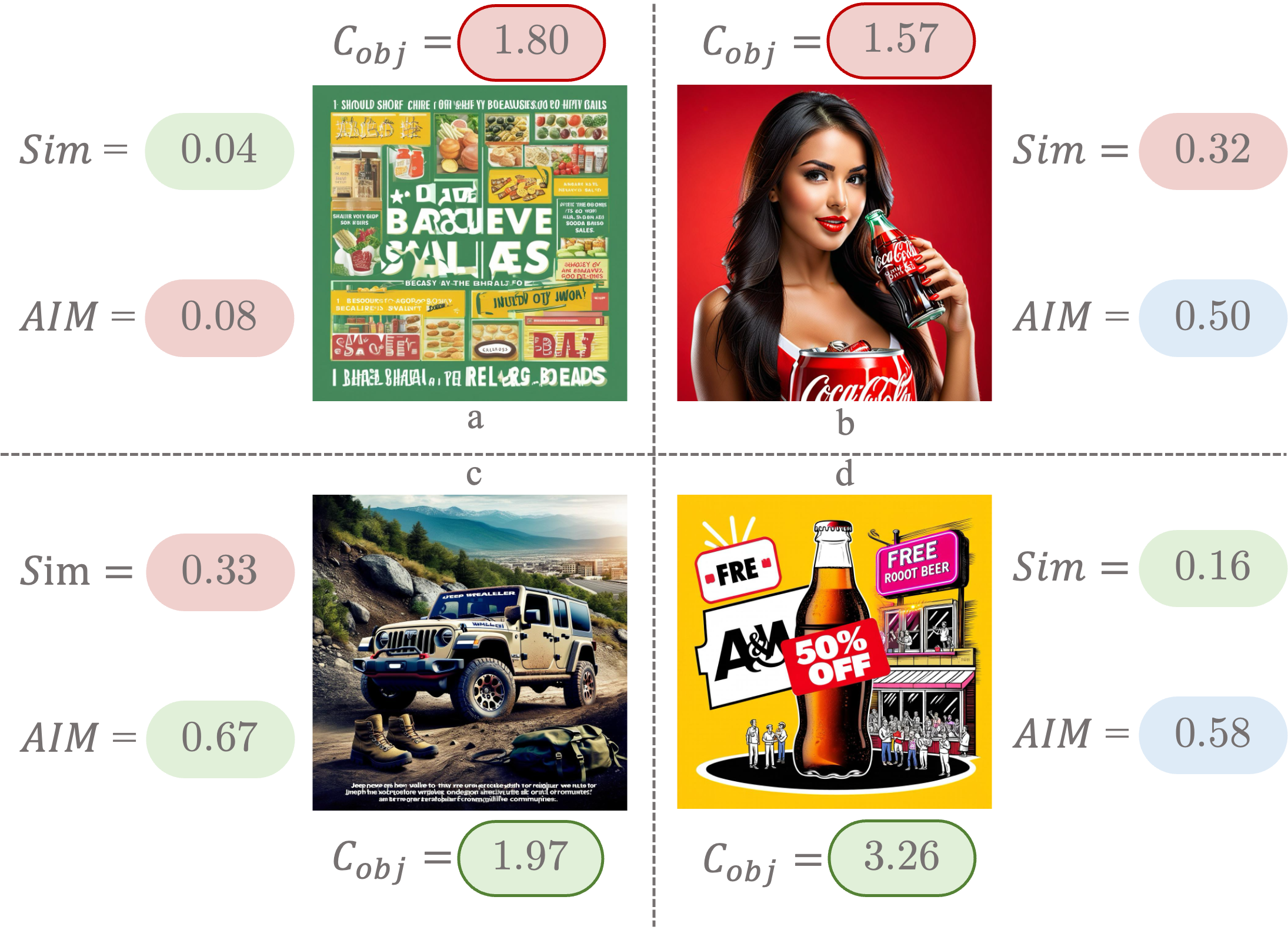}
    \vspace{-0.2cm}
    \caption{Creativity:
    \textbf{(d)} which shows unique but relevant object portrayal scores best, while random \textbf{(a)} and generic objects \textbf{(b)} score worst. \textcolor{Red}{Red}/\textcolor{Green}{Green} shows \textcolor{Red}{low}/\textcolor{Green}{high} values for $C_{obj}$, \textcolor{Red}{low}/\textcolor{Green}{high} for $AIM$, \& \textcolor{Red}{high}/\textcolor{Green}{low} for $Sim$. 
    Blue denotes moderate values.}
    \label{fig:rebuttal}
\end{figure}

\textbf{Qualitative example.} Fig.~\ref{fig:rebuttal} shows examples of $C_{obj}$ and its components AIM and \textit{Sim}.
Fig.~\ref{fig:rebuttal}\textbf{a} was generated based on $AR_m$ ``I should shop here because they have good sales'' (objects: sales, foods, deals). It shows unusual objects (good \textit{Sim} score) which are however irrelevant (bad AIM score), hence bad creativity overall. 
Fig.~\ref{fig:rebuttal}\textbf{b} (``I should drink Coca-Cola because attractive women do'', obj: Coca-Cola) achieves the lowest score (moderate AIM, typical objects).
Fig.~\ref{fig:rebuttal}\textbf{c}
has high alignment with the prompt, but the objects are not creative. Fig.~\ref{fig:rebuttal}\textbf{d} (``I should drink this root bear because it has a discount''; obj: root bear, A$\&$W drink) has the highest overall score due to unusual portrayal of the objects and good alignment with the prompt.  

\begin{table}[!tp]
\centering
\scriptsize
\setlength{\tabcolsep}{0.8pt}
\begin{tabular}{l|c|c|c}
    \toprule  
    Annotators & COM & PSA  & All \\
    \midrule
    \midrule
    H, $C_{LLM}$ & -0.03 & 0.15 & 0.04 \\ 
    H, $C_{obj}$ & \textbf{0.57} & \textbf{0.53} &  \textbf{0.54} \\
    \hline
    H1, H2 & 0.70 & 0.78  & 0.73 \\
\bottomrule
\end{tabular}
\caption{Agreement among human annotators, $C_{obj}$ and baseline. 
}
\label{tab:creativity_agreement}
\end{table}


\textbf{Reliability of the proposed Creativity score.} 
As observed in Table~\ref{tab:creativity_agreement}, the Krippendorff's Alpha agreement between human annotators and the baseline metric $C_{LLM}$ is -0.03 for Commercial advertisements, indicating systematic disagreement \cite{agreement}, and 0.15 for PSA advertisements, suggesting a low and close to random level of agreement \cite{agreement}. Our proposed metric, $C_{obj}$, shows significant improvement over these baseline measures, achieving agreement of 0.57 for Commercial and 0.53 for PSA advertisements, representing moderate agreement \cite{agreement}. 
While $C_{obj}$ does not yet reach human-human agreement levels, it significantly decreases the gap, offering a reliable and robust metric for creativity evaluation.
In supp, we include results from baseline 
metrics on both real and generated ads. 

\textbf{T2I models lack creativity.} 
In Table \ref{tab:creativity_alignment}, we observe $I_{LLM}$ achieves higher creativity scores than $I_{AR}$ in all cases, showcasing the effectiveness of using $\mathcal{E}_{LLM}$ expansions. QwenLM achieves significantly better results than LLAMA3 as the source of the expansion. Among LLMs for evaluation (AIM), LLAMA3 has a slight edge for commercial ads, while other scores are similar.

\begin{table}
\centering
\scriptsize
\setlength{\tabcolsep}{0.8pt}
\begin{tabular}{l|c|c|c}
    \toprule  
    Annotators & COM & PSA  & All \\
    \midrule
    \midrule
    H, $P_{LLM}$ & 0.27 & 0.26 & 0.27 \\
   H, $P_{comp}$ (InternVL, LLAMA3) & 0.83 & 0.54 & 0.65 \\
   H, $P_{comp+AIM}$ (InternVL, LLAMA3) & 0.85 &\textbf{ 0.75} &\textbf{ 0.80} \\ 
    H, $P_{comp+AIM}$ (QwenVL, LLAMA3) & 0.73 & 0.63 & 0.68\\ 
    H, $P_{comp+AIM}$ (InternVL, QwenLM) & \textbf{0.89} & 0.30 & 0.63\\ 
    H, $P_{comp+AIM}$ (QwenVL, QwenLM) & \textbf{0.89} & 0.74 & 0.74\\ 
    \hline
    H1, H2 & 0.80 &  0.56 & 0.70 \\
\bottomrule
\end{tabular}
\caption{Agreement with human annotators for different persuasiveness scores.  
We show different combinations of MLLMs and LLMs for $P_{comp+AIM}$ by ([MLLM], [LLM]). 
}
\label{tab:p_agreement}
\end{table}




\subsection{Persuasiveness}

\textbf{Accuracy of $P_{comp+AIM}$.} Due to the lack of baseline metrics for persuasiveness, we compare our proposed metric with $P_{LLM}$, in Table~\ref{tab:p_agreement}. All $P_{comp}$ and $P_{comp+AIM}$ results outperform $P_{LLM}$, confirming the value of our components-based persuasion evaluation. Further, the difference between $P_{comp+AIM}$ and $P_{comp}$ shows the value of adding alignment to the components. 

\begin{table}[!htp]
\centering
\scriptsize
\setlength{\tabcolsep}{0.8pt}

\begin{tabular}{l|c|c|c|c|c|c|c|c}
    \toprule  
    Annotators& E & S & O & I & AU & B & AP & All\\
    \midrule
    \midrule
    H, $P_{comp+AIM}$* & -0.15 & -0.03 & 0.24 & 0.06 & 0.21 & 0.05 & 0.25 & 0.78\\
    \hline
    H1, H2 & 0.74 & 0.40 & 0.74 & 0.40 & 0.53 & 0.54 & 0.34 & 0.89\\

\bottomrule
\end{tabular}
\caption{Agreement among annotators and LLM responses for each persuasiveness component (column headers). ``All'' compares the image chosen most frequently across the components by humans, and the image with higher average score from $P_{comp+AIM}$.}
\label{tab:agreement_persuasiveness}
\end{table}

\begin{table}[!htp]
\centering
\setlength{\tabcolsep}{0.8pt}
\scriptsize
\begin{tabular}{l|c|c|c|c|c|c|c|c|c}
    \toprule  
    Ad type & $I_{input-text}$ & T2I  & E & S & O & I & AU & B & AP  \\
    \midrule
    \midrule
    \multirow{4}{*}{COM} & \multirow{2}{*}{$I_{AR}$} & SDXL & 0.74 & 0.66 & 0.54 & 0.63 & 0.69  & 0.58 & 0.67 \\
    & & AuraFlow &  0.74 & 0.70 & \textbf{0.69} & 0.65 & 0.68  & 0.57 & 0.66 \\
    
    & $I_{LLM}$ & AuraFlow & \textbf{0.77} & \textbf{0.72} & 0.63 & \textbf{0.72} & \textbf{0.72} & \textbf{0.67} & \textbf{0.71} \\
    \cline{2-10} 
    & \multicolumn{2}{c|}{$I_{real}$} & 0.74 &  {0.76} & 0.66 & 0.65 & {0.74}  & {0.67} & {0.71} \\
    \midrule
    \midrule
    \multirow{4}{*}{PSA} & \multirow{2}{*}{$I_{AR}$} & SDXL &  0.59  & 0.53 & 0.52 & 0.49 & 0.53 & 0.23 & 0.55 \\
    & & AuraFlow &  0.57 & 0.48 & \textbf{0.75} & 0.52 & 0.63 & 0.34 & 0.52 \\
    
    & $I_{LLM}$ & AuraFlow & \textbf{0.71} & \textbf{0.70} & {0.68} & \textbf{0.70} & \textbf{0.67} & \textbf{0.37} & \textbf{0.65} \\
    \cline{2-10}
    & \multicolumn{2}{c|}{$I_{real}$} & 0.66 & {0.71}  & {0.68} & 0.56 & 0.62 & {0.39} & {0.65} \\
\bottomrule
\end{tabular}
\caption{Scoring direct and expansion-based T2I generation (and real ads) using different components of persuasiveness.}
\label{tab:com_persuasiveness}
\end{table}

\textbf{Accuracy of different components of persuasiveness.} Table~\ref{tab:agreement_persuasiveness} represents the agreement among human annotators and scores from $P_{comp}$ across each component used in the $P_{comp}$ score (Fig.~\ref{fig:PA}). When comparing the agreement among human annotators across all images, Tab.~\ref{tab:p_agreement} and \ref{tab:agreement_persuasiveness} show higher agreement when we directly ask the annotator to choose the more persuasive image (0.80, Tab.~\ref{tab:p_agreement}) compared to agreement with combined per-component responses (0.78 under ``All'', Tab.~\ref{tab:agreement_persuasiveness}).
This suggests that the individual component questions are more subjective than directly asking the human annotators to choose the more persuasive images. However, average human agreement over these persuasiveness components (Tab.~\ref{tab:agreement_persuasiveness}) is still reasonable: 0.53 average of Tab.~\ref{tab:agreement_persuasiveness} columns except ``All'', is interpreted as moderate agreement, and the combined questions (``All'') agreement is 0.89, showing the potential for using these questions. 
Importantly, using these components in formulating the $P_{comp+AIM}$ prediction of the AI system is also more helpful than directly asking the LLM to score persuasiveness ($P_{LLM}$, Tab.~\ref{tab:p_agreement}). 

\textbf{Low persuasion in images generated by T2I models.} As observed in Table~\ref{tab:creativity_alignment}, the best $P_{comp+AIM}$ scores are always achieved using LLM expansion. Evaluation using QwenVL as the MLLM is slightly worse than InternVL. Among T2I methods, FLUX is worse than AuraFlow (top part of the table).
Table~\ref{tab:com_persuasiveness} compares different images across each component of persuasiveness. $I_{LLM}$ (using LLAMA3) generated by Auraflow outperforms $I_{AR}$ also generated by Auraflow in 6 out of 7 components in both Commercial and PSA advertisements. This further highlights the effectiveness of our proposed approach in generating more persuasive images.

\section{Conclusion}
\label{sec:conclusion}
We examined the challenges of generating and evaluating persuasive and creative advertisements from visually implicit messages. We introduced three metrics—Creativity, Alignment, and Persuasiveness (CAP)—that achieved high agreement with human annotations. We showed existing T2I models struggle to generate images that are creative, aligned, and persuasive. We improved CAP scores through a simple yet effective approach using LLM-based expansion of abstract messages requiring interpretation of implicit content, and used these as prompts to T2I models.

\textbf{Acknowledgment.} This work was partly supported by NSF Grant No. 2006885 and partly by the University of Pittsburgh Center for Research Computing and Data, RRID:SCR\_022735, through the resources provided. Specifically, this work used the H2P cluster, which is supported by NSF award number OAC-2117681. We gratefully acknowledge the support of our annotators.

{   \small
    \bibliographystyle{ieeenat_fullname}
    \bibliography{main}     }

\maketitlesupplementary
In this work, we introduced three new evaluation metrics to assess: 1. alignment of images with implicit messages, 2. creativity, and 3. persuasiveness. In the supplementary material, we provide:

\begin{enumerate}
    \item Implementation details.
    \item A detailed table for text-image alignment, including metrics from prior works, such as ImageReward, VQAScore and CLIPScore (Table \ref{tab:alignment_full}).
    \item Results from baseline metrics for evaluating creativity and persuasiveness (Table \ref{tab:persuasion_creativity}).
    \item Two examples showing the benefit of combining the persuasiveness components for better alignment with human evaluations.
    \item A comparison of PA components with human evaluations for persuasiveness (Table \ref{tab:agreement_p_real_images}) on real advertisements ($I_{real})$.
    \item Example showing impact of different $\alpha$ values.
    \item Examples of failures and discussion of limitations.
    \item The prompts used for MLLMs and LLMs.
    \item Human annotation interface.
\end{enumerate}

 \section{Implementation Details}

We fine-tuned LLAMA3-Instruct-8B (LLAMA3) and Qwen-2.5-7B-Instruct (QwenLM) LoRA \cite{LORA} parameters, for generating accurate action-reason statements for each image ($AR_{gen}$), using the CPO trainer \cite{CPO}, with a batch size of 4, learning rate of 5e-5, and 3,000 training steps. 
We sampled 250 images, utilizing each of the three to five $AR_m$ messages from the PittAd dataset \cite{PittAd} as accepted statements in the training data. Each accepted statement was paired with all 4 hard negatives for the corresponding image from \cite{wacv}, to create negative statements. This process created a training set of 3,500 data points. 

For LLAMA3-Instruct-8B \cite{dubey2024llama}, Qwen-2.5-7B-Instruct \cite{QWenLM}, QwenVL-2-7B-Instruct \cite{QWenVL}, and InternVL-2-26B \cite{InternVL}, AuraFlow-2 \cite{AuraFlow} and FLUX.1-dev \cite{Flux} we used the default temperature settings with 8-bit quantization. For SDXL \cite{sdxl} and FLUX.1-dev \cite{Flux}, we applied the default guidance scale, no negative prompt, and 28 inference steps. For AuraFlow-2 \cite{AuraFlow}, we set a guidance scale of 5 and used 28 inference steps. 

All experiments were conducted on A100 and L40s GPUs.

\section{Alignment Results}

\begin{table}[!htp]
\centering
\scriptsize
\setlength{\tabcolsep}{0.9pt}
\begin{tabular}{l|l|c|c|c|c|c|c|c|c}
\hline
\multicolumn{2}{c|}{\textbf{Image}} & \multicolumn{4}{c|}{\textbf{COM Ads}} & \multicolumn{4}{c}{\textbf{PSA Ads}} \\ \cline{1-10} 
\textbf{$I_{input-text}$} &  \textbf{T2I} & \textbf{AIM} & \textbf{IR} & \textbf{VQA} & \textbf{CS} & \textbf{AIM} & \textbf{IR} & \textbf{VQA} & \textbf{CS}\\ \hline 
\hline
\multirow{2}{*}{$I_{AR}$} &SDXL& 0.50 & \textbf{0.44} & 0.72 & 0.23 & 0.32  &  \textbf{-0.05}  & \textbf{0.75} & 0.24\\ 
& AuraFlow  & 0.50 & 0.00 & 0.75 & 0.22 & 0.31 & -0.21 & 0.72 & 0.24\\

$I_{LLM}$  & AuraFlow & {0.53} & 0.29 & \textbf{0.76} & 0.23 & {0.43}  & -0.13 & 0.69  & 0.23\\ \hline

\multicolumn{2}{c|}{$I_{real}$} & \textbf{0.55 }& -0.69 & 0.75 & \textbf{0.26} & \textbf{0.49}  &  -1.03 & 0.72 & \textbf{0.26}\\ \hline
\end{tabular}
\caption{Comparison of alignment of the images with action-reason message ($AR_m$) using our proposed AIM and baseline metrics. Best result per column bolded (including real images). IR = ImageReward. CS = CLIPScore. This result uses InternVL and LLAMA3.}
\label{tab:alignment_full}
\end{table}

Table~\ref{tab:alignment_full} represents the text-image alignment scores for different images and evaluated by AIM and baseline metrics. 
Since $I_{real}$ are the ground-truth images in the dataset and $AR_m$ are the interpretation of real images, we expect $I_{real}$ to have higher alignment score than $I_{gen}$ (i.e. $I_{AR}$, $I_{LLM}$). 
However, the table highlights that ImageReward (IR) \cite{ImageReward} and VQAScore (VQA) \cite{VQA} assign lower scores to $I_{real}$ (ground-truth) compared to generated images. 
This contradicts our expectation and highlights the limitations of these metrics when evaluating the alignment of images and implicit prompts.
CLIPScore (CS) \cite{CLIP_score} performs reasonably as it rates $I_{real}$ the highest, but it rates the alignment of $I_{LLM}$ and $I_{AR}$ equally. This is problematic because human annotators preferred $I_{LLM}$ in 92\% of comparisons, demonstrating its superior alignment. AIM assigns the highest score to $I_{real}$, followed by $I_{LLM}$ which highlights the higher accuracy of our proposed evaluation method. 

\section{Creativity and Persuasiveness Baseline Metrics}

\begin{table}[!tp]
\centering
\scriptsize
\setlength{\tabcolsep}{0.9pt}
\begin{tabular}{l|l|c|c|c|c|c|c|c|c}
\hline
\multicolumn{2}{c|}{\textbf{Image}} & \multicolumn{4}{c|}{\textbf{COM Ads}} & \multicolumn{4}{c}{\textbf{PSA Ads}} \\ \cline{1-10} 
\textbf{$I_{input-text}$} &  \textbf{T2I} & \textbf{$C_{obj}$} & \textbf{$C_{LLM}$} & \textbf{$P_{c+A}$} & \textbf{$P_{LLM}$} & \textbf{$C_{obj}$} & \textbf{$C_{LLM}$} & \textbf{$P_{c+A}$} & \textbf{$P_{LLM}$}  \\ \hline 
\hline
\multirow{2}{*}{$I_{AR}$} & SDXL & 2.03 & 0.63 & 0.62 & 0.60 & 1.33 & 0.59 & 0.48 & 0.52\\ 
& AuraFlow  & 2.12 & 0.73 & 0.64 & 0.60 & 1.36 & \textbf{0.72} & 0.42 & 0.57\\
$I_{LLM}$  & AuraFlow  &\textbf{ 2.25} & \textbf{0.77} &\textbf{ 0.70} & \textbf{0.76} & \textbf{1.87} & 0.71 & \textbf{0.60} & \textbf{0.67}\\ \hline

\multicolumn{2}{c|}{$I_{real}$} & {2.28} & 0.67 & {0.98} & 0.66 & {2.04} & 0.65 & 0.60 & 0.61\\ \hline
\end{tabular}
\caption{Comparison of persuasiveness and creativity proposed ($C_{obj}$, $P_{c+A}$) and baseline metrics. Best gen. image result bolded.}
\label{tab:persuasion_creativity}
\end{table}

$I_{real}$ are images from PittAd dataset \cite{PittAd} designed to be creative and persuasive and are expected to be more creative than $I_{gen}$. As observed in Table~\ref{tab:persuasion_creativity} the $C_{LLM}$ score for $I_{real}$, which represents the ground-truth images, is lower than the scores for both $I_{LLM}$ and $I_{AR}$ generated by AuraFlow-2, by 10\% and 6\% respectively, across both Commercial and PSA advertisements. This raises questions about the reliability of LLMs in accurately scoring creativity, as real images receive lower scores than generated ones. In contrast, our proposed metrics assign the highest scores to $I_{real}$, demonstrating a more consistent alignment with human expectations.

\section{Examples of Benefit of Combining Persuasiveness Components }

\begin{figure}[h]
    \centering    \includegraphics[width=1\linewidth]{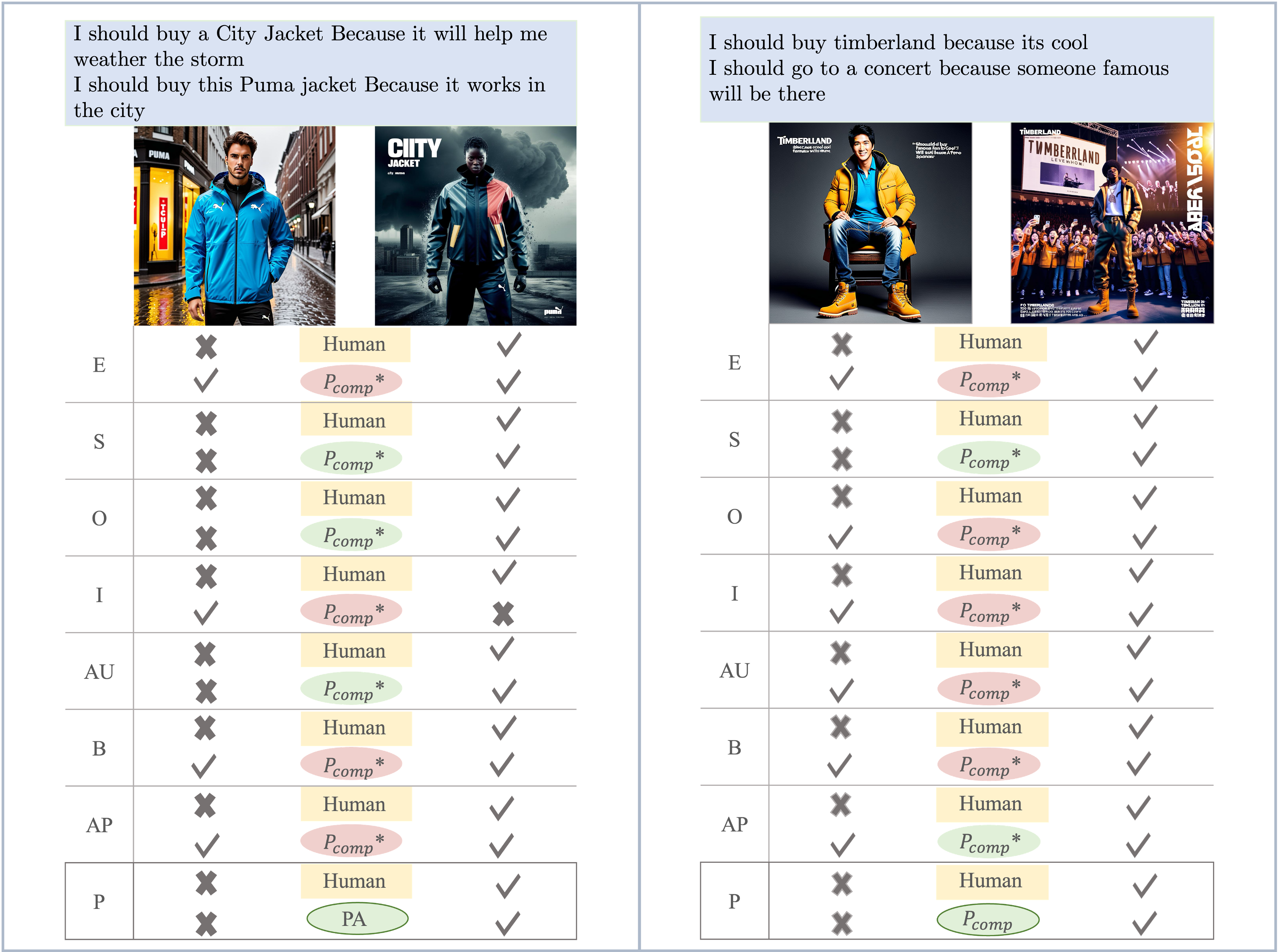}
    \caption{Example of images chosen by human annotators for each component in $P_{comp}$ score compared to $P_{comp}$ score. In each row, a $\checkmark$ represents the chosen image for the corresponding component by human or the persuasiveness component, while a $\times$ indicates the rejected image. \textcolor{Green}{Green} represents the agreement of score with human annotator while \textcolor{Red}{red} represents disagreement.}
    \label{fig:PAexample}
\end{figure}

\section{Persuasiveness Metrics on Real Ads}

Due to the inherent differences in quality between $I_{real}$ and $I_{gen}$, we avoided direct comparisons in human evaluations to reduce bias in selecting the better option. Instead, we compared two $I_{real}$ images from the same topic. Since the images were distinct, their corresponding $AR_m$ messages also differed, making alignment comparisons non-applicable. However, we evaluated the persuasiveness components for the two $I_{real}$ images and reported the agreement levels between the annotators, as well as between the annotators and the LLM, across all components. The results are presented in Table~\ref{tab:agreement_p_real_images}. We also analyze the agreement between the most frequently chosen images by annotators across all components and the average scores assigned to each image for these components. As observed in Table~\ref{tab:agreement_p_real_images}, while the agreement for individual components is lower than chance, combining all components results in an average agreement increase of 0.44. It is worth noting that the agreement between the $P_{comp+AIM}*$ metric and human annotators is higher than the agreement between the human annotators.

\begin{table}[!htp]
\centering
\scriptsize
\setlength{\tabcolsep}{0.8pt}

\begin{tabular}{l|c|c|c|c|c|c|c|c}
    \toprule  
    Annotators& E & S & O & I & AU & B & AP & All\\
    \midrule
    \midrule
    H, $P_{comp+AIM}$* & -0.53 & -0.27 & 0.03 & 0.22 & 0.0 & -0.34 & -0.18 & 0.34\\
    \hline
    H1, H2 & -0.22 & 0.31 & 0.19 & 0.38 & -0.60 & 0.38 & -0.17 & 0.31\\

\bottomrule
\end{tabular}
\caption{On real ads, agreement among annotators and the LLM responses for each persuasiveness component (column headers). ``All'' 
computes agreement across all components: for H, we get the image chosen most frequently across all the components, and for $P_{comp+AIM}$, we choose the image with higher average score.}
\label{tab:agreement_p_real_images}
\end{table}


\subsection{Impact of Different $\alpha$ Values}

\begin{figure}
    \centering
    \includegraphics[width=0.75\linewidth]{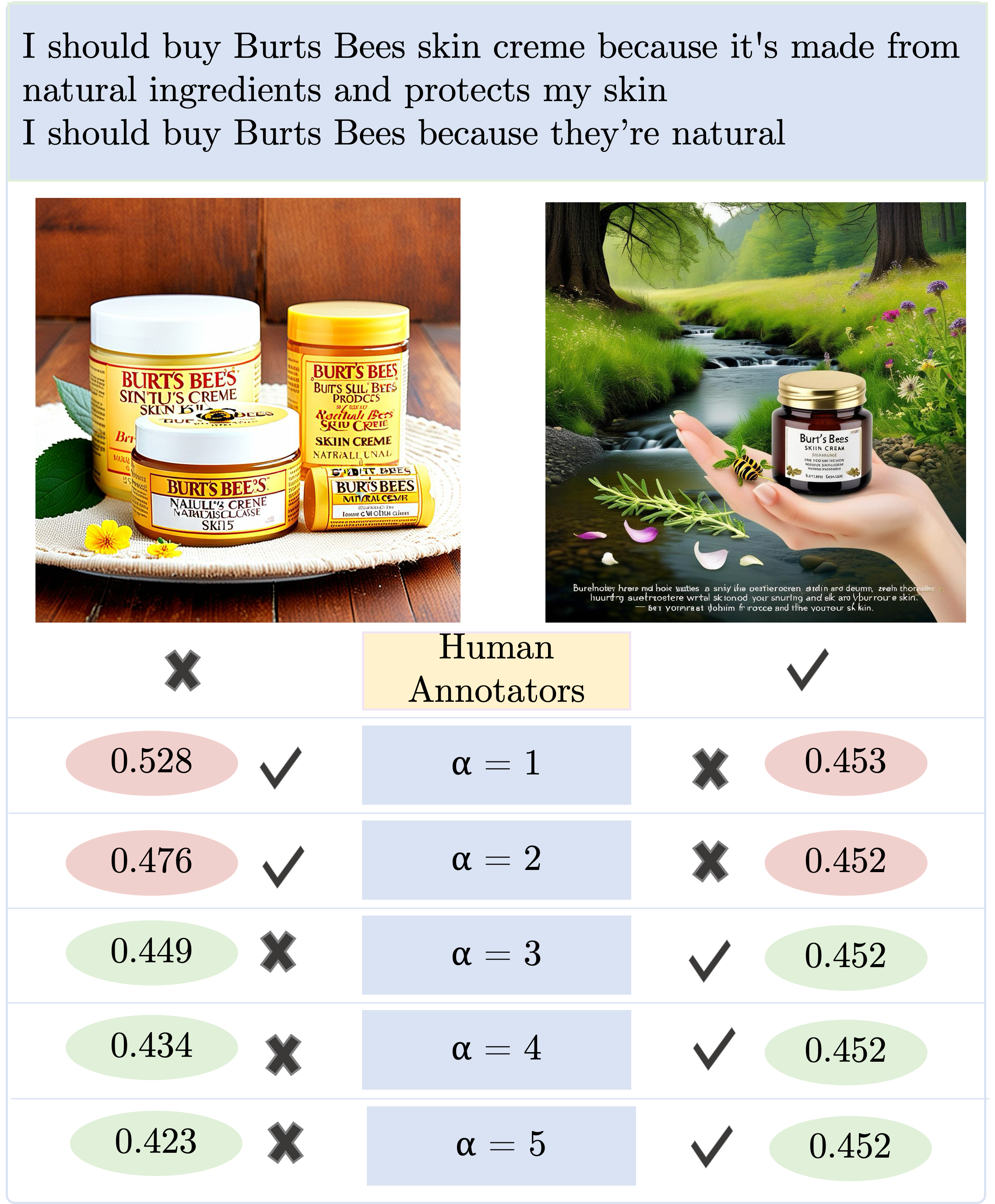}
    \caption{Example of different values for alpha in the AIM score. In each row, the value under each image indicates the score generated by the AIM with the alpha listed for that row. A $\checkmark$ shows the chosen image, and a $\times$ indicates the rejected image. \textcolor{Green}{Green} shows agreement with the human in choosing the more aligned image and \textcolor{Red}{red} highlights the disagreement with the human.}
    \label{fig:alpha_example}
\end{figure}

Fig. \ref{fig:alpha_example} shows an example of AIM scores with different $\alpha$, in Eq. 1 in the main paper. Results are similar for values of $\alpha$ greater than one (i.e. greater impact of alignment over the reason in the ad message).

\section{Failure Examples and Limitations}

\begin{figure}[!htp]
    \centering
    \includegraphics[width=1\linewidth]{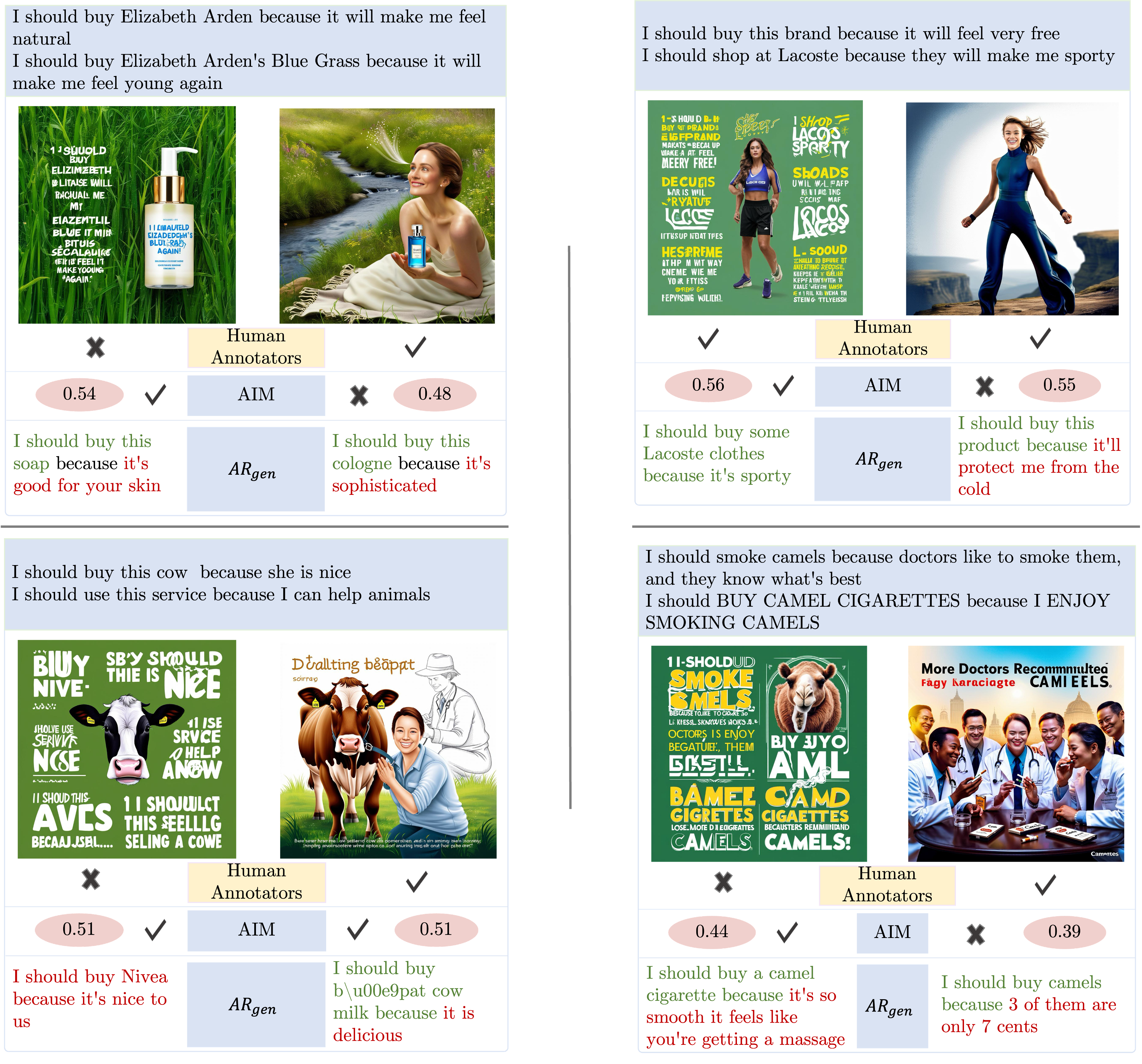}
    \caption{Example of failure of AIM in both generating accurate $AR_{gen}$ and choosing the more aligned image with $AR_m$. \textcolor{Red}{Red text} indicates inaccurate action/reason considering the image. \textcolor{Green}{Green text} shows accurate action/reason. \textcolor{red}{Red circle} shows inaccurate choice of image for which image is more aligned with the $AR_m$ according to the human annotation.}
    \label{fig:failure}
\end{figure}

In this section, we first highlight examples of failures in the AIM method, specifically in generating accurate $AR_{gen}$ and selecting images that align with human annotators' choices, as illustrated in Fig.~\ref{fig:failure}. Next we discuss the limitation of our proposed metrics.

\subsection{Failure examples}

As illustrated in Fig.~\ref{fig:failure}, top-right example, when the image fails to convey the intended message accurately, the model may hallucinate while generating $AR_{gen}$. For instance, in this example, the image depicts wind, which can imply coldness, but it does not effectively represent the clothing's ability to keep someone warm. Due to the unclear intent of the image, the LLM struggles to interpret the message correctly and hallucinates while generating the action-reason statement. Additionally, when the MLLM incorporates textual elements into the image description (as seen in the left image of each example in Fig.~\ref{fig:failure}), the fine-tuned LLM often generates an $AR_{gen}$ that closely resembles the intended $AR_m$, even if the \emph{visual} content is misaligned. However, due to the unclear or incorrect spelling in the text, human annotators do not consider these images aligned with the statement. To address this issue, we force the MLLM to ignore the texts that starts with ``I should" or ``I shouldn't" (which are common starting words for action-reason statements). Additionally, we prompt the MLLM to determine whether the image is text-only. If the image is identified as text-only, a score of 0 is returned to account for the lack of meaningful visual content.

\subsection{Limitation}
Given the focus of this work, we train and evaluate our methods exclusively on the PittAd dataset \cite{PittAd}. For the evaluation of persuasiveness, we employ an LLM to address seven questions in a single evaluation. While this approach is less efficient, it ensures comprehensive assessment, as the subjective nature of the components makes it essential to include all components for an accurate evaluation of persuasiveness.

\section{Prompts}
In this section, we present the prompts used in different evaluation methods. We begin with the prompts for generating descriptions, as shown in Listings~\ref{prompt:train_description} and \ref{prompt:inference_description}. Next, we provide the prompts for generating action-reason statements based on image descriptions in Listings~\ref{prompt:alignment_zeroshot} and \ref{prompt:alignment_ft}. Then, we present the prompts used for evaluating the creativity and persuasiveness of images with LLMs, detailed in Listings~\ref{prompt:creativity} and \ref{prompt:persuasiveness}. Finally, we outline the prompts for evaluating individual $P_{comp}$ components, including elaboration, synthesis, originality, imagination, audience targeting, benefit conversion, and appeal category, in Listings~\ref{prompt:elaboration}, \ref{prompt:synthesis}, \ref{prompt:originality}, \ref{prompt:imagination}, \ref{prompt:audience}, \ref{prompt:benefit}, and \ref{prompt:appeal}.
\begin{lstlisting}[caption={Description generation for images using InternVL-v2-26B\cite{InternVL} prompt templates in generating train data.}, label={prompt:train_description}]
Describe the image in detail in one paragraph.
Only return the description. Do not include any further explanation.
\end{lstlisting}

\begin{lstlisting}[caption={Description generation for images with MLLMs prompt templates for inference. If the answer to Q1 is ``No", then the AIM score for that image is 0.}, label={prompt:inference_description}]
Carefully analyze the image and respond only in the specified format, without any interpretations or inferences. Focus on only the visible elements in the image. Ensure that any object seen in the image is included in Q1, even if it is described in more detail in Q2.

Response Format:

Q1: ${answer to Q1}
Q2: ${answer to Q2}

Questions:

Q1: Are there any objects in the image, excluding text-only logos, and text? List at most 5 such objects if present.
Q2: Describe the image in detail, focusing only on visible objects and elements without adding any interpretation, opinion, or analysis.
\end{lstlisting}
\begin{lstlisting}[caption={Prompt for Action-Reason Prediction with zero-shot LLAMA3 \cite{dubey2024llama} and fine-tuned QwenLM \cite{QWenLM}. \textcolor{highlight-blue}{\{Blue\}} denotes elements added dynamically, based on the description of the image.}, label={prompt:alignment_zeroshot}]
What is the correct interpretation
for the described image:
Description: (*@\textcolor{highlight-blue}{\{description\}}@*).
The interpretation format is: I should ${action} because ${reason}. ONLY RETURN A SINGLE SENTENCE IN THIS FORMAT
\end{lstlisting}
\begin{lstlisting}[caption={Prompt for Action-Reason Prediction with fine-tuned LLM\cite{dubey2024llama} in AIM evaluation. \textcolor{highlight-blue}{\{Blue\}} denotes elements added dynamically, based on the description of the image.}, label={prompt:alignment_ft}]
What is the correct interpretation for the described image:
Description: (*@\textcolor{highlight-blue}{\{description\}}@*).
\end{lstlisting}
\begin{lstlisting}[caption={Prompt for evaluation of creativity with LLMs. \textcolor{highlight-blue}{\{Blue\}} denotes elements added dynamically, based on the description of the image.}, label={prompt:creativity}]
(*@\textbf{USER:}@*)
Description: (*@\textcolor{highlight-blue}{\{description\}}@*)
Question: Score the creativity of the described image in range of (0, 5).
Your output format is only Answer: score\n form, no other form. Empty is not allowed.
(*@\textbf{ASSISTANT:}@*)
\end{lstlisting}
\begin{lstlisting}[caption={Prompt for evaluation of persuasiveness with LLAMA3-Instruct\cite{dubey2024llama}. \textcolor{highlight-blue}{\{Blue\}} denotes elements added dynamically, based on the description of the image.}, label={prompt:persuasiveness}]
(*@\textbf{USER:}@*)
(*@\textbf{Context:}@*) If the image convinces the audience to take an action, it is considered persuasive.
Description: (*@\textcolor{highlight-blue}{\{description\}}@*)
Question: Based on the context score the persuasiveness of the described image in the range of (0, 5).
Your output format is only Answer: score\n form, no other form. Empty is not allowed.
(*@\textbf{ASSISTANT:}@*)
\end{lstlisting}

\begin{lstlisting}[caption={Prompt for scoring elaboration with LLMs. \textcolor{highlight-blue}{\{Blue\}} denotes elements added dynamically.}, label={prompt:elaboration}]
(*@\textbf{Context:}@*) You are supposed to score the image based on the descriptions.
You are given the description of an image and you are asked to score the image based on the following question.
Description: (*@\textcolor{highlight-blue}{\{description\}}@*)
How visually detailed is the image? Do not consider the text in the image. Return a score in the range (0,5).
Please follow the format of:
    Explanation: ${explanation}
    Answer: ${answer}
(*@\textbf{ASSISTANT:}@*)
\end{lstlisting}

\begin{lstlisting}[caption={Prompt for scoring synthesis with LLMs. \textcolor{highlight-blue}{\{Blue\}} denotes elements added dynamically.}, label={prompt:synthesis}]
(*@\textbf{Context:}@*) You are supposed to score the image based on the descriptions.
You are given the description of an image and you are asked to score the image based on the following question.
Description: (*@\textcolor{highlight-blue}{\{description\}}@*)
How well does the image connect the objects that are usually unrelated? Return the score in range (0,5).
Please follow the format of:
    Explanation: ${explanation}
    Answer: ${answer}
(*@\textbf{ASSISTANT:}@*)
\end{lstlisting}

\begin{lstlisting}[caption={Prompt for scoring originality with LLM. \textcolor{highlight-blue}{\{Blue\}} denotes elements added dynamically.}, label={prompt:originality}]
(*@\textbf{Context:}@*) You are supposed to score the image based on the descriptions.
You are given the description of an image and you are asked to score the image based on the following question.
Description: (*@\textcolor{highlight-blue}{\{description\}}@*)
How out of the ordinary, and unique the image is, and how well it breaks away from habit-bound and stereotypical thinking? Return a score in range (0, 5).
Please follow the format of:
    Explanation: ${explanation}
    Answer: ${answer}
(*@\textbf{ASSISTANT:}@*)
\end{lstlisting}



\begin{lstlisting}[caption={Prompt for scoring imagination with LLMs. \textcolor{highlight-blue}{\{Blue\}} denotes elements added dynamically.}, label={prompt:imagination}]
(*@\textbf{Context:}@*) You are supposed to score the image based on the descriptions.
You are given the description of an image and you are asked to score the image based on the following question.
Description: (*@\textcolor{highlight-blue}{\{description\}}@*)
Assume you are a human evaluating the given image. How well does the image allow you to form images you have not directly experienced before more easily? Do not consider the text in the image. Return a score in the range (0,5)
Please follow the format of:
    Explanation: ${explanation}
    Answer: ${answer}
(*@\textbf{ASSISTANT:}@*)
\end{lstlisting}

\begin{lstlisting}[caption={Prompt for scoring how well the image targets audience with LLMs. \textcolor{highlight-blue}{\{Blue\}} denotes elements added dynamically.}, label={prompt:audience}]
(*@\textbf{Context:}@*) You are supposed to score the image based on the descriptions.
You are given the description of an image and you are ask to score the image based on the following question.
Description: (*@\textcolor{highlight-blue}{\{description\}}@*)
How well the image targets (*@\textcolor{highlight-blue}{\{audience\}}@*)? Return a score in range (0, 5)
Please follow the format of:
    Explanation: ${explanation}
    Answer: ${answer}
(*@\textbf{ASSISTANT:}@*)
\end{lstlisting}

\begin{lstlisting}[caption={Prompt for scoring how well the image converts features to benefits for costumer with LLMs. \textcolor{highlight-blue}{\{Blue\}} denotes elements added dynamically.}, label={prompt:benefit}]
(*@\textbf{Context:}@*) You are supposed to score the image based on the descriptions.
You are given the description of an image and you are asked to score the image based on the following question.
Description: (*@\textcolor{highlight-blue}{\{description\}}@*)
How well does the image connect the feature of the products to the benefits for customers? Return the score in range (0,5).
Please follow the format of:
    Explanation: ${explanation}
    Answer: ${answer}
(*@\textbf{ASSISTANT:}@*)
\end{lstlisting}


\begin{lstlisting}[caption={Prompt for scoring how well the image targets correct appeal category with LLMs. \textcolor{highlight-blue}{\{Blue\}} denotes elements added dynamically.}, label={prompt:appeal}]
(*@\textbf{Context:}@*)You are supposed to score the image based on the descriptions.
You are given the description of an image and you are asked to score the image based on the following question.
Description: (*@\textcolor{highlight-blue}{\{description\}}@*)
To answer the question consider the following explanation:
    There are three types of rhetorical appeals or ways to convince the audience:

        - Ethos is a persuasive technique that appeals to an audience by highlighting credibility. Ethos advertisement techniques invoke the superior character of a speaker, presenter, writer, or brand.
        - Pathos is a persuasive technique that tries to convince an audience through emotions. Pathos advertisement techniques appeal to the senses, memory, nostalgia, or shared experience.
        - Logos is the persuasive technique that aims to convince an audience by using logic and reason. Also called the logical appeal, logos examples in advertisements include the citation of statistics, facts, charts, and graphs.

Question: Assume you are a human evaluating the image. Based on the context score, how well the image appeals to (*@\textcolor{highlight-blue}{\{appeal\-category\}}@*) in the range of (0,5).
Please follow the format of:
    Explanation: ${explanation}
    Answer: ${answer}
(*@\textbf{ASSISTANT:}@*)
\end{lstlisting}


\end{document}